%% file: main.tex
\documentclass[runningheads,dvipsnames]{llncs}

\usepackage[T1]{fontenc}
\usepackage{graphicx}
\usepackage{booktabs}

\usepackage{cite}
\usepackage{amsmath,amssymb,amsfonts}
\usepackage{algorithmic}
\usepackage{textcomp}
\usepackage{cleveref}
\usepackage{float}
\usepackage{xcolor}
\usepackage{xspace}
\usepackage{mathtools}

\usepackage{placeins}
\usepackage{comment}
\usepackage{suffix,xstring}
\usepackage[nolist]{acronym}

\newcommand{\larr}{\leftarrow}
\newcommand{\rarr}{\rightarrow}

\newcommand{\Rarr}{\Rightarrow}
\newcommand{\infin}{\infty}

\newcommand{\LG}{LRP-$\gamma$\xspace}
\newcommand{\SG}{SmoothGrad\xspace}

\usepackage{pifont}
\newcommand{\cmark}{\ding{51}}%
\newcommand{\xmark}{\ding{55}}%

\begin{document}
\begin{acronym}
    \acro{ml}[ML]{Machine Learning}
\end{acronym}
\begin{acronym}
    \acro{svd}[SVD]{Singular Value Decomposition}
\end{acronym}
\begin{acronym}
    \acro{lrp}[LRP]{Layer-wise Relevance Propagation}
\end{acronym}
\begin{acronym}
    \acro{sg}[SG]{SmoothGrad}
\end{acronym}
\begin{acronym}
    \acro{ig}[IG]{Integrated Gradient}
\end{acronym}
\begin{acronym}
    \acro{shap}[Shapley]{Shapley value method}
\end{acronym}
\begin{acronym}
    \acro{gi}[G$\times$I]{Gradient$\times$Input}
\end{acronym}

\title{Uncovering the Structure of Explanation Quality with Spectral Analysis}



%
%
\author{Johannes Maeß\inst{1,2}\orcidID{0009-0002-0792-4348}
\and \\
Grégoire Montavon\inst{1,3}\orcidID{0000-0001-7243-6186} 
\and \\
Shinichi Nakajima\inst{1, 2,6}\orcidID{0000-0003-3970-4569} 
\and \\
Klaus-Robert Müller\inst{1,2,4,5}\orcidID{0000-0002-3861-7685}
\and \\
Thomas Schnake\inst{1,2}\orcidID{0009-0006-3768-0259} }
\authorrunning{Maeß et al.}
%
\institute{
BIFOLD -- Berlin Institute for the Foundations of Learning and Data, Berlin, Germany
\and
Machine Learning Group, TU Berlin, Berlin, Germany\\
\email{\{maess,t.schnake,nakajima,klaus-robert.mueller\}@tu-berlin.de}
\and
Charit\'e -- Universit\"atsmedizin Berlin, Germany\\
\email{gregoire.montavon@charite.de}
\and
Department of Artificial Intelligence, Korea University, Seoul, Korea
\and
Max-Planck Institute for Informatics, Saarbr{\"u}cken, Germany
\and
RIKEN AIP, Tokyo, Japan
}

\sloppy

\maketitle              
\input{chapters/00abstract}
\input{chapters/01introduction}
\input{chapters/03theory}
\input{chapters/05empirical}
\input{chapters/07conclusionandfuture}

\section*{Acknowledgment}
This work was in part supported by the Federal
German Ministry for Education and Research (BMBF) under Grants
BIFOLD24B, BIFOLD25B, 01IS14013A-E, 01GQ1115, 01GQ0850, and 031L0207D.
KRM was partly supported by the Institute of Information \& communications Technology Planning \& Evaluation (IITP) grants funded
by the Korea government (MSIT) (No. RS-2019-II190079, Artificial
Intelligence Graduate School Program, Korea University and No. RS2024-00457882, AI Research Hub Project). We thank Stefan Blücher and Adrian Hill for suggestions of experiments and valuable discussions. We thank Nicole Trappe for proofreading of the manuscript.

\appendix
\input{chapters/11appendix}
\bibliographystyle{splncs04}
\bibliography{references}
\end{document}

%% file: chapters/00abstract.tex
\begin{abstract}
As machine learning models are increasingly considered for high-stakes domains, 
effective explanation methods are crucial to ensure that their prediction strategies are transparent to the user. Over the years, numerous metrics have been proposed to assess quality of explanations. However, their practical applicability remains unclear, in particular due to a limited understanding of which specific aspects each metric rewards. In this paper we propose a new framework based on spectral analysis of explanation outcomes to systematically capture the multifaceted properties of different explanation techniques. Our analysis uncovers two distinct factors of explanation quality-\textit{stability} and \textit{target sensitivity}--that can be directly observed through spectral decomposition. Experiments on both MNIST and ImageNet show that popular evaluation techniques  (e.g., pixel-flipping, entropy) partially capture the trade-offs between these factors. Overall, our framework provides a foundational basis for understanding explanation quality, guiding the development of more reliable 
techniques for evaluating explanations.

\keywords{
Explainable AI \and 
Spectral Analysis \and 
Stability \and 
Sensitivity \and 
Interpretability \and 
Explanation Quality \and 
Transparency \and 
Machine Learning.
}

\end{abstract}

%% file: chapters/01introduction.tex
\section{Introduction}
\label{sec:introduction}
Machine learning (ML) models are being adopted in critical domains such as healthcare, finance, and media, often surpassing human capabilities in various tasks. However, their internal decision-making processes--particularly in artificial neural networks--are notoriously opaque, effectively making them `black boxes'. As these models permeate high-stakes domains, the need for robust transparency and interpretability has grown. This lack of transparency has spurred the development of Explainable AI (XAI), a field that focuses on extracting explanations for the predictions of complex ML models.

A range of explanation methods have been proposed (see, e.g., \cite{baehrens2010explain, strumbelj2014explaining, caruana2015healthcare, bach2015pixel, koh2017influence, xiong22asubgraph,xiong23bMaxProd,schnake2022gnnlrp,schnake2025symbxai}), each of which has been shown to be useful in specific domains. Yet, with the large number of XAI methods, users face the challenge of selecting an appropriate method and assessing its quality. Additionally, evaluating the quality of XAI methods is essential not only for improving human interpretability but also because heat maps have been reported to mislead users, a concern heightened by their recent use in sensitive domain such as healthcare \cite{hense2025xmilinsightfulexplanationsmultiple} and industry \cite{letzgus2023xaiframeworkrobusttransparent}. For instance, adversarial attacks can manipulate explanation methods \cite{dombrowski2019explanations}, and models may base their predictions on spurious correlations rather than meaningful patterns, a phenomenon known as the `Clever-Hans' effect \cite{lapuschkin2019unmasking}.

Previous work \cite{swartout1993expert,nauta2023xai_eval_review} highlights that evaluating an explanation's quality is inherently multifaceted. While numerous evaluation techniques have been proposed (e.g.,\ \cite{samek2016evaluating,dabkowski2017locate,adebayo2018sanity}) to assess explanation quality, there is still limited clarity about which specific aspects each technique rewards. 
There has been longstanding effort to holistically assess explanations (e.g.,\ \cite{swartout1993expert,hedstrom2023quantus}), yet this remains largely empirical. In contrast, we aim to develop a deeper theoretical understanding of how evaluation techniques are interconnected.  

In this paper, we seek a better theoretical understanding of the question of explanation quality. We propose a novel spectral analysis approach that applies \ac{svd} to a matrix containing input-neuron contributions for each output neuron, thereby characterizing explanations via their singular values. We demonstrate that these singular values can reveal whether an explanation meets specific desired properties: Sensitivity to the model's output (termed \emph{target sensitivity}) or distinct identification of salient features (termed \emph{stability}).

Our spectral analysis is tested on two image datasets, MNIST and ImageNet. Specifically, we investigate how the explanation techniques and their hyperparameters relate to the sensitivity and stability properties identified in our spectral analysis. Our analysis also allows us to check whether parameters that perform well on both properties are consistent with those identified by pixel-flipping and other commonly used evaluation techniques. Our results provide further support for existing approaches for evaluating explanations.

In summary, our work unifies previously disparate explanation quality metrics and proposes a framework to guide the development of more robust evaluation techniques.

\section{Related work}

In this section, we review two areas of XAI that are most related to the problem studied here, namely, the problem of evaluation, and the spectral analyses that operate on explanation techniques. For a broader discussion of XAI and its applications, we refer to the review papers \cite{arrieta2020review, samek2021explaining, klauschen2024pathology}.

\subsection{Evaluation of Explainable AI Methods}\label{sec:relate_work_eval_method}

Evaluating XAI methods has become a pressing concern, prompting the development of a wide range of approaches (cf.\ \cite{nauta2023xai_eval_review} for a review). In the context of expert systems, \cite{swartout1993expert} identifies a number of desiderata aimed at holistically characterizing what constitutes a good explanation. Other research efforts have focused on mathematically defining what makes an explanation desirable, establishing specific axioms or unit tests that any effective explanation should satisfy \cite{shapley1953axioms, adebayo2018sanity, montavon2019gradient, binder2023shortcomings}.

Another category of work addresses the question of evaluating explanations by performing direct tests on the model, specifically testing whether removing features deemed relevant by the explanation results in a substantial change in the output of the model. These methods include pixel-flipping and its many variants \cite{bach2015pixel, samek2016evaluating, petsiuk2018rise, ancona2018attribution, bluecher2024decoupling}. Pixel-flipping tracks the change in the target prediction as individual pixels are removed (flipped) in order of relevance, from most to least relevant. Denoting $i_1, \dots, i_{d}$ the feature indices sorted by relevance, and $x_{\{i_1, \dots, i_k\}}$ a data point where the $k$ most relevant features have been replaced with a placeholder value, a `pixel-flipping curve' $\{\phi(x_{\{i_1, \dots, i_k\}})\}_{k = 1}^{d}$ is created, where $\phi$ is the ML-model. The smaller the area under the pixel-flipping curve (PF-AUC), the more successful the explanation was at identifying the truly relevant features.

Another category of methods evaluates the consistency of an explanation against established  `ground-truth' knowledge, such as leveraging the outputs of a high-performing vision model. Several studies have proposed verifying whether the spatial distribution of relevance scores aligns with the regions corresponding to visual objects detected by the model \cite{dabkowski2017locate,zhang2018locate}. 

In \cite{yu2017interpret_shannon_regular} and \cite{tseng2020xai_shannon_noise}, the authors suggest using \emph{Shannon entropy} to assess whether the explanation is free of highly entropic noise patterns, which is a prerequisite for these explanations to be faithful and interpretable. The entropy can be quantified by first converting the explanation into a probability vector $\overline{R} = |R| / 1^\top |R|$ and then calculating the entropy as $ -\sum_{i}  \overline{R}_{i} \cdot \log (\overline{R}_{i})$.  

Efforts have been made to develop benchmarks and software tools to enable a comprehensive, multifaceted evaluation of explanation techniques \cite{hedstrom2023quantus}. Finally, \cite{doshivelez2017rigorous, selvaraju2020gradcam} describes end-to-end evaluation settings where a human recipient is actively involved and where performance metrics can be more easily defined.

\subsection{Spectral Methods for Explainable AI}

Spectral methods have interacted with the field of XAI in several ways. Spectral Relevance Analysis (SpRAy) \cite{lapuschkin2019unmasking} generates a collection of explanations from a dataset and uses spectral clustering to identify clusters of prototypical decision strategies. While both SpRAy and our approach rely on singular values, our approach differs by focusing on characterizing the explanation for a single data point, rather than analyzing the broader decision strategies of an entire model.

Spectral methods are also employed in \cite{chormai2024disentangled}, where eigenvalues of a cross-covariance matrix, which links model activations to responses, are used to assess the complexity of a decision strategy. While the techniques used are related, our work focuses on the distinct problem of evaluating an explanation.  

The authors of \cite{sixt2020explanations} analyze the rank of explanations in propagation-based methods and observe a rank collapse under certain conditions, causing explanations to lose their task specificity. While it shares the same general goal as our work of evaluating and better understanding explanation techniques, our work differs in that it looks at the globality of the explanation spectrum and makes connections to other properties of explanations, such as their stability.

%% file: chapters/03theory.tex
\section{Spectral Analysis of Explanation Quality}
\label{section:method}

In this section, we present our spectral analysis framework for analyzing model explanations. Its purpose is to gain better insight into the structure of explanation quality, by uncovering underlying factors that contribute to it.

Our analysis will apply to ML models, typically neural networks, that map some input vector $x \in \mathbb{R}^d$ to some output $z \in \mathbb{R}^h$ via some function $\phi:\mathbb{R}^d \to \mathbb{R}^h$ learned from the data. The model's output may consist of $h$ class logits or an abstract $h$-dimensional representation, suitable for linear readouts. We focus on attribution-based explanation techniques, which assign a relevance score to each input feature, indicating its importance in the model's prediction.

Our framework begins by encoding the attribution from each output to each input in a $d \times h$ redistribution matrix $R_{\cdot | \cdot}$. Each column is defined as:
$$
R_{\cdot | j} = \frac{\mathcal{E}(z_j)}{1^\top \mathcal{E}(z_j)}
$$
for all $j=1\dots h$. Here $\mathcal{E}(z_j) \in \mathbb{R}^d$ denotes the attribution of the model's output $z_j$ at class $j$ to the input features, given by the explanation method of interest at one specific data point. The matrix $R_{\cdot | \cdot}$ satisfies the property $1^\top R_{\cdot|\cdot} = 1$. In particular, when its entries are non-negative, it acquires a probabilistic interpretation, where its elements correspond to the percentage of each output that is redistributed to each input of the network.

This redistribution matrix is particularly useful because it enables any quantity $y \in \mathbb{R}^h$ at the network output to be propagated backward via matrix-vector multiplication $\mathcal{E}(y) = R_{\cdot|\cdot} y$, producing the desired explanation. For example, if the output of the network represents class logits, we can explain evidence for class $j$ by defining $y = e_j \odot z$ where $e_j$ is a one-hot vector of the $j$th dimension, and then matrix multiplying by $R_{\cdot|\cdot}$. Likewise, defining $y = (e_j - e_{j'}) \odot z$ enables to explain the log-likelihood ratio between two classes. In a general case, when the output of the network is an abstract representation, defining $y = w \odot z$ enables to explain the readout with the weights $w$.

We now would like to characterize general properties of the explanation process in a way that is independent on the exact prediction task. In particular, we analyze the inherent behavior of the multiplication of $y$ by $R_{\cdot|\cdot}$, and whether it amplifies or attenuates the magnitude of the vector $y$ it multiplies to. These amplification/attenuation properties can be characterized by the spectrum of $R_{\cdot | \cdot}$, that is, the collection of singular values extracted by singular value decomposition (SVD):
\begin{align}
R_{\cdot | \cdot}\stackrel{\text{(SVD)}}{=}\ \sum_{i=1}^K \sigma_i u_i v_i^\top
\label{eq:svd}
\end{align}
with $K = \min(d,h)$ and $\sigma_1 \geq \sigma_2 \geq \dots \geq \sigma_K \geq 0$ are the singular values. The largest singular value $\sigma_1$ is equivalent to the spectral norm of the redistribution matrix $\|R_{\cdot|\cdot}\|_2$ and corresponds to the maximum amplification that can be experienced during the explanation process. In turn, $\sigma_2, \sigma_3, \dots$ represent second, third, etc. largest amplification factors along orthogonal directions $u_2, u_3, \dots$ and $v_2, v_3, \dots$ in both input and output space, respectively.

\subsection{Stability of an Explanation}

The \emph{stability} of an explanation technique is defined by its ability to produce consistent explanations that remain unaffected by factors irrelevant to the model's prediction strategy. One manifestation of irrelevant factors in the context of deep neural networks is the shattered gradient effect \cite{balduzzi2017shattered}, where strong variations of the ML-model manifest themselves only locally. Simple gradient-based explanation techniques such as Gradient$\,\times\,$Input (see \cite{ancona2019gradient}) are strongly affected by them and tend to produce noisy explanations that perform poorly on explanation benchmarks (e.g.,\ \cite{binder2023shortcomings}).

Thus, the content of an explanation should be limited to what is strictly necessary to support the prediction $y$. Assuming our explanations $\mathcal{E}(y)$ satisfy the conservation property ($1^\top \mathcal{E}(y) = 1^\top y$), such an objective can be enforced with a small norm $\|\mathcal{E}(y)\|$, where $\|\cdot\|$ represents any operator norm. If $\|\mathcal{E}(y)\|$ is small, we can generally assume that the heat map is rather smooth or stable.  

Our spectral analysis is especially useful to quantify this property as it gives an upper bound on the norm of the explanation in terms of the first singular value $\sigma_1$ of the redistribution matrix $R_{\cdot | \cdot}$, i.e.:
\begin{align}  \label{eq:regularity}
\frac{\|\mathcal{E}(y)\|_2}{\|y\|_2} \leq \sigma_1
\end{align}
In other words, it is desirable for an explanation technique to have a low $\sigma_1$. Similar relations to Eq.\ \eqref{eq:regularity} can be stated beyond the spectral norm $\sigma_1 =\|R_{\cdot|\cdot}\|_2$ for other norms of $\|R_{\cdot|\cdot}\|_p$. We show in Section \ref{sec:formal_lrp}, in the context of the \ac{lrp} explanation method \cite{bach2015pixel, montavon2019layer}, how $\|R_{\cdot|\cdot}\|_1$ (and the resulting explanation \emph{stability}) can be controlled by an appropriate choice of the \ac{lrp} parameter $\gamma$.

It should be noted, however, that a stable explanation like the aforementioned is not sufficient to ensure high quality explanations. A uniform redistribution matrix of the type $R_{\cdot | \cdot} = \mathbf{1}/d$, has a very small $\sigma_1$, but this results in a uniform redistribution over the input features, which is undesirable since it lacks any \emph{sensitivity} to the output, and thus fails to discriminate between features contributing to different predicted outputs.

\subsection{Sensitivity of an Explanation}
\begin{figure}[b!]
\centering
\includegraphics[width=.85\textwidth]{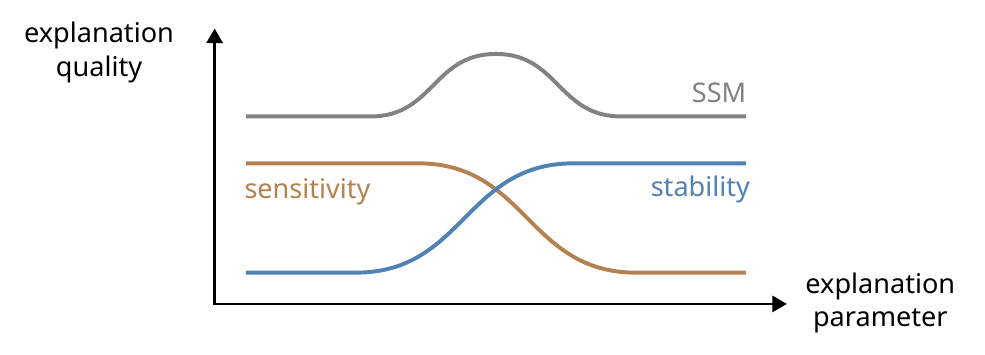}
\caption{Cartoon depiction of the two factors of explanation quality that can be derived from our spectral analysis and the SSM metric (Eq.\ \eqref{eq:cartoon}) that aggregates them. We postulate the existence of a `sweet spot' where both explanation \emph{stability} and \emph{sensitivity} can be achieved.
This can be reached by a subtle adjustment of explanation parameters such as LRP's $\gamma$ and SmoothGrad's standard deviation (as later shown in Figures \ref{fig:comparison_vgg16_g16_all} and \ref{fig:comparison_d3_g14_all}).
} \label{fig:cartoon}
\end{figure}

Sensitivity of the explanation to different outputs, in other words, the ability to distinguish attributions between classes, is an important additional property of an explanation, and its importance has been highlighted in several works (e.g., \cite{adebayo2018sanity, binder2023shortcomings, nauta2023xai_eval_review}).   To illustrate how \emph{sensitivity} can be related to the spectral properties of the explanation, let us consider two different output neurons $j$ and $j'$, accessed by the readout functions $y = e_j \odot z$ and $y' = e_{j'} \odot z$. If the two output neurons encode different concepts, and the latter are supported by different input features, it is fair to say that the two explanations should be different. Using our spectral analysis, we can express these two explanations and decompose them in terms of the singular values extracted by our spectral analysis: 
\begin{align}
\mathcal{E}(y)
&= R_{\cdot|\cdot} \, y\phantom{'} = \textstyle \sum_i \sigma_i u_i v_i^\top y\\
\mathcal{E}(y')
&= R_{\cdot|\cdot} \, y' = \textstyle \sum_i \sigma_i u_i v_i^\top y'
\end{align}
A key observation in these equations is that the explanations of $y$ and $y'$ are tightly controlled by the spectra of the network. In the extreme case, when $\sigma_2 = \dots = \sigma_K = 0$, $\mathcal{E}(y)$ and $\mathcal{E}(y')$ become mere rescalings of each other--the rapid decay of the singular value spectrum severely limits the capacity to generate explanations that accurately capture the diverse concepts present at the output. This phenomenon has also been observed in \cite{sixt2020explanations} for certain backpropagation-based techniques in deep networks. Maintaining large values for all singular values, which can be quantified by the norm $\|(\sigma_k)_{k=1}^K\|_2$, is, therefore, essential to ensure the \emph{sensitivity} of explanations.

\subsection{Stability-Sensitivity Metric}

Having singled out $\sigma_1$, or more precisely its inverse $1/\sigma_1$, as a factor of explanation \emph{stability}, and $\|(\sigma_k)_{k=1}^K\|_2$ as a factor of explanation \emph{sensitivity}, it comes quite natural to combine them into a single score, which we define as the `\emph{stability}-\emph{sensitivity} metric' (SSM):
\begin{align} \label{eq:cartoon}
    \text{SSM} = \frac{1}{\sigma_1} \cdot \|(\sigma_k)_{k=1}^K\|_2
\end{align}
The SSM and the two factors on which it depends are sketched in Fig.\ \ref{fig:cartoon}
for a hypothetical explanation parameter that interpolates between a stable but insensitive explanation and a sensitive but unstable explanation. This explanation quality metric and the two factors that make it up will be compared to existing evaluation metrics such as pixel-flipping in later experiments. The computational complexity of evaluating the \emph{stability}, \emph{sensitivity}, and the \emph{SSM} is discussed in Appendix \ref{sec:computational_complexity}.

\subsection{Linking LRP Parameters to Stability and Sensitivity}\label{sec:formal_lrp}

To demonstrate how the parameters of an explanation technique can influence the \emph{stability}-\emph{sensitivity} profile of an explanation method, we perform a theoretical analysis for the case of the \ac{lrp} \cite{bach2015pixel, montavon2019layer} explanation technique. LRP is an explanation technique that operates by propagating the output of a neural network layer-by-layer to the input features, where each propagation step consists of the application of a purposely designed propagation rule. Extending the framework of Section \ref{section:method}, we can characterize the LRP explanation process as a composition of multiple redistribution steps. In other words, it is achieved through a sequence of multiplications with redistribution matrices defined at each layer:
\begin{align}
R_{\cdot|\cdot}^{\text{LRP}} = R_{\cdot|\cdot}^{(1)} \cdot R_{\cdot|\cdot}^{(2)} \cdot \dots \cdot R_{\cdot|\cdot}^{(L)}
\label{eq:lrpchain}
\end{align}
We specifically consider the case where the propagation rule LRP-$\gamma$ (cf.\ \cite{montavon2019gradient}) is applied at each layer. At a given layer, let $j$ and $k$ denote the indices of input and output neurons, respectively. The LRP-$\gamma$ rule then defines the resulting redistribution scheme:
\begin{align}
R_{j|k}^{(l)} :=  \frac{a_{j} \cdot (w_{jk} + \gamma w_{jk}^+)}{\sum_{j'} a_{j'} \cdot (w_{j'k} + \gamma w_{j'k}^+)}
\label{eq:lrprule}
\end{align}
where $a_j$ and $a_k$ are the neuron activations and $w_{jk}$ are the weights connecting these neurons. The parameter $\gamma$ emphasizes positive contributions, which is instrumental in controlling the explanation behavior of LRP.

Next, to analyze the \emph{stability} of $R_{\cdot|\cdot}^\text{LRP}$, we combine matrix norm identities and the form of Eq.\ \eqref{eq:lrpchain}, to derive the following chain of inequalities:
\begin{align}
\frac{\|\mathcal{E}(y)\|_p}{\|y\|_p} \leq  \|R_{ \cdot | \cdot}^\text{LRP}\|_p \leq \prod_{i=1}^L \|R_{\cdot|\cdot}^{(l)}\|_p
\end{align}
which hold for any $p$. In the case of $p=2$ the expression reduces to an expression involving spectral norms computed either globally or individually for each layer. If, instead, we choose $p=1$, we get a closed-form expression where the layer-wise terms can be written analytically as $\| R_{\cdot|\cdot}^{(l)}\|_1 = 1 + 2c^{(l)}/(1-c^{(l)} + \gamma)$ with $c^{(l)} = \max_{k} |\sum_j [h_j w_{jk}]^-| / \sum_{j}[h_j w_{jk}]^+ \in [0,1)$ (the proof is given in the Appendix). This leads to the closed-form relation:
\begin{align} \label{eq:l1_operator_norm_network}
\frac{\|\mathcal{E}(y)\|_1}{\|y\|_1}
&\leq \prod_{l=1}^L\left(1 + \frac{2c^{(l)}}{1-c^{(l)} + \gamma}\right)
\end{align}
This shows that increasing $\gamma$ tightens the bound on the operator norm, promoting \emph{stability}. As observed in prior work (e.g.,\ \cite{pahde2023optimizing, chormai2024disentangled}), small $\gamma$ values result in noisy explanations and poor benchmark performance. Notably, both noise reduction and explanation \emph{stability} are achieved rapidly, as Eq.~\eqref{eq:l1_operator_norm_network} saturates relatively quickly with increasing $\gamma$.

%% file: chapters/05empirical.tex
\section{Experiments}
\label{sec:empirical}

In this section, we empirically evaluate our proposed XAI framework on neural networks trained for the MNIST \cite{lecun2010mnist} and ImageNet \cite{deng2009imagenet} vision tasks, using four different explanation methods. We then compare our results to established metrics, highlighting the strengths and insights offered by our spectral analysis approach.

\subsection{Experimental Setup}
\paragraph{Machine Learning Models.}
We evaluate our approach on two established datasets of different sizes and characteristics.
MNIST, containing gray-valued 28$\times$28 images for the task of digit classification, and ImageNet, with colored images of size 224$\times$224 picturing everyday objects of 1000 classes.
For ImageNet, we download a pre-trained VGG16 model, specifically the \texttt{IMAGENET1K\_V1} weights from the torchvision library \cite{marcel2010torchvision}. The model achieves a Top-5 accuracy of 90\% on images sampled from the ILSVRC 2012 validation set \cite{deng2009imagenet}, meaning in 90\% of the the predictions the correct class is within the most probable 5 class predictions. For MNIST, we use a Convolutional Neural Network (CNN) (11 layers; Top-1 accuracy 98\%) whose artificially deep architecture makes it challenging to explain. For more information on the training and architecture of the model on MNIST we refer to Appendix \ref{sec:mnist_cnn_arch}.

\paragraph{Explanation Techniques}  
We employ multiple explanation techniques to compare our framework against existing evaluation techniques. 
Below, we detail how we apply and parameterize the different explanation techniques.  

\begin{enumerate}  
\setlength{\itemsep}{1.mm}
    \item[] \textbf{Layer-wise Relevance Propagation (\ac{lrp})}: For the \ac{lrp} explanation method \cite{bach2015pixel}, we follow the approach of \cite{montavon2019layer}. Specifically, on the VGG-16 ImageNet model, we use the \LG rule in the first three blocks of the architecture, and the $z^\mathcal{B}$-rule in the first layer.
    For the small MNIST-CNN, we use \LG for all layers except the last one. In our experiments, we analyze the effect of varying $\gamma$ on the different measures of explanation quality. 

    \item[] \textbf{\ac{sg}}: For \ac{sg} \cite{smilkov2017smoothgrad}, we calculate the gradient of the model 10 times, each time under a zero-mean Gaussian noise perturbation of the input. In our experiments, we analyze the impact of different smoothing parameters $s$ (i.e.\ standard deviation values for the noise).

    \item[] \textbf{\ac{ig}}: In the explanation method \ac{ig}\cite{sundararajan2017axiomatic}, we approximate the integral by 10 equidistant evaluations of the model. The reference point in the integral is a black image.  

    \item[] \textbf{Shapley Value Sampling (Shapley)}: We use the Shapley method \cite{trumbelj2010AnEE} as an additional baseline,
    and limit the number of feature removal cycles to 25. The scheme we use for removing pixels and patches consists of filling them with uniform black color. 
\end{enumerate}

\paragraph{XAI Evaluation Metrics}

We compare our spectral analysis framework with the evaluation methods pixel-flipping \cite{bach2015pixel,samek2016evaluating} and Shannon entropy.
In pixel-flipping, we remove increasing sets of the most relevant features. We stop after 5\% of the total features are flipped, and calculate the area under the curve (PF-AUC) as a summary of how faithful the explanation is to the model. The lower the PF-AUC, the better the truly relevant features have been identified. 
Placeholder values for flipped pixels in the input image are in-painted using the OpenCV \cite{opencv_library} implementation of the Fast-Marching algorithm \cite{telea2004image}, utilizing 3 and 5 pixels around the deleted section for MNIST and ImageNet inputs, respectively.

We then compare our spectral analysis to Shannon entropy, which is commonly used to detect noise and evaluate the overall readability of heat maps. We calculate the Shannon entropy according to the formula given in Section \ref{sec:relate_work_eval_method}. Explanations are evaluated on 100 different input images each.

\begin{figure}[t]
    \centering 
    \includegraphics[width=0.9993\textwidth]{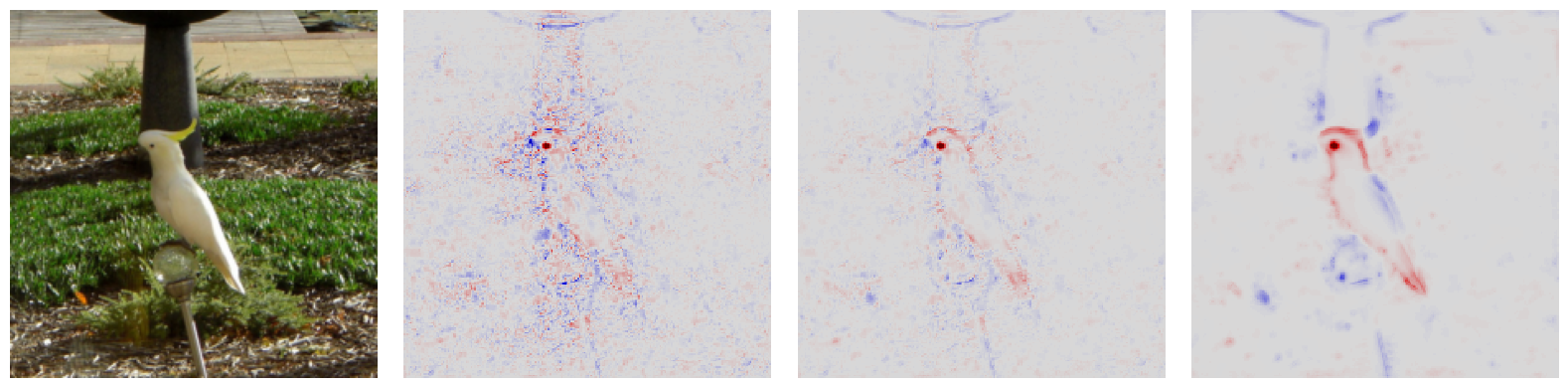}
    \hspace{-5mm}
    \begin{tabular}{cccc}  
&
         \parbox{2.9cm}{\centering $\gamma=0.005$} &
         \parbox{2.9cm}{\centering $\gamma=0.04$} &
         \parbox{2.9cm}{\centering $\gamma=0.3$} \\\midrule
        \parbox{2.9cm}{\centering Stability} & \xmark & \cmark & \cmark \\
        \parbox{2.9cm}{\centering Sensitivity} & \cmark & \cmark & \xmark \\
    \end{tabular}
    \caption{Examples of explanations produced by the LRP explanation technique using the rule LRP-$\gamma$ with different values of the parameter $\gamma$. An increase in $\gamma$ is associated with an increase in explanation \emph{stability} (visible here as the vanishing noise pattern in the explanation). On the other hand, choosing too large a value for $\gamma$ results in a decrease in target \emph{sensitivity}.}
\label{fig:vogel_hms}
\end{figure}

\subsection{Comparison of XAI Evaluation Methods with Spectral Analysis}
We first provide an qualitative description of heat maps that fulfill \textit{stability} or \textit{sensitivity} to build an intuition of how either manifest visually, and then show a quantitative analysis of the different properties we are interested in on a variety of explanation methods.

\begin{figure}[ht!]
    \centering
    \includegraphics[width=0.999\textwidth]{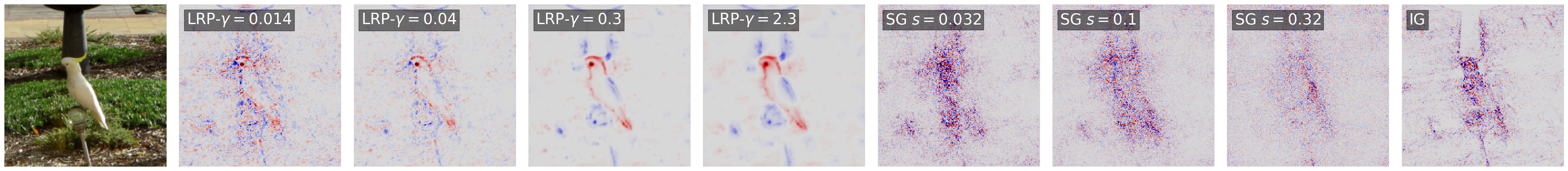}
    \includegraphics[width=0.999\textwidth]{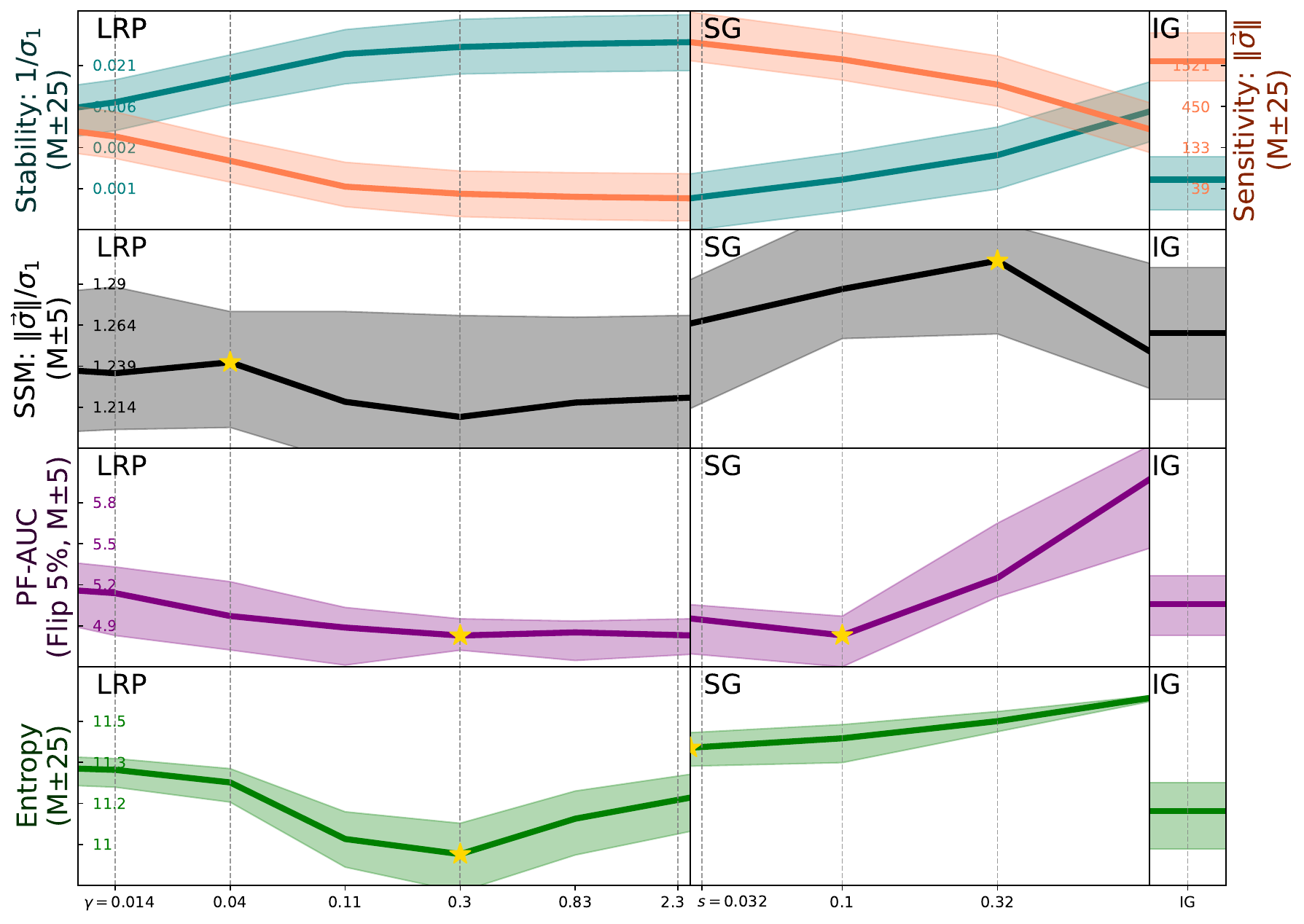}
    \caption{Quantitative analysis of evaluation metrics for explanation methods on an ImageNet-trained model using 100 images. The top row shows exemplary heat maps for the class `sulphur-crested cockatoo’, one per method and parameter choice per dashed vertical line drawn in the main plot. Below, we present evaluation metrics (top to bottom): \emph{stability} \& \emph{sensitivity} (ours), SSM (ours), PF-AUC, and entropy. Explanation methods (left to right) include LRP, \ac{sg}, and \ac{ig}. For LRP, results for 11 $\gamma$ values are shown, and for \ac{sg}, 4 different noise levels. PF-AUCs are calculated after deleting 5\% of the image. Thick lines indicate the median, with shaded areas showing variability: 5\% for SSM and PF-AUC, 25\% for \emph{stability}, \emph{sensitivity}, and entropy. The star indicates where explanation quality under the given metric is maximized.
}
    \label{fig:comparison_vgg16_g16_all}
\end{figure}

\subsubsection{Qualitative Assessment of Stability and Sensitivity}

In Fig. \ref{fig:vogel_hms}, we see how the $\gamma$ parameter in the LRP method influences the heat maps and how each heat map corresponds to specific properties, such as \emph{stability} and \emph{sensitivity}. 
For a small $\gamma$ parameter (i.e., $\gamma = 0.005$), \emph{stability} is low but the \emph{sensitivity} is high. This means that the explanation may vary with small output perturbation, yet is distinguishable from different classes. Effectively, we see that the heat map looks relatively noisy and irregular, which aligns with our expectation, since a small $\sigma_1^{-1}$ does not limit the growth of noise in the explanation phase.
The other extreme is when $\gamma$ is high (i.e., $\gamma =0.3$), where we observe high \emph{stability} but low \emph{sensitivity}. The absence of noise in the explanation gives it high \emph{stability}. However, its low \emph{sensitivity} suggests that heat maps may not be able to distinguish features specific to each class, thereby reducing its overall usefulness as an explanation.
For the case when $\gamma=0.04$, a `sweet spot' is reached where both \emph{stability} and \emph{sensitivity} are high. The noise remains small and the heat map is easy to interpret, while the \emph{sensitivity} measure indicates that heat maps are specific to each class.

 \subsubsection{Quantitative Comparison with Different Evaluation Methods}
We consider in our quantitative analysis the MNIST and ImageNet models described above, and apply to each model different explanation techniques and evaluations of these explanations. In addition to the LRP and \ac{sg} explanation techniques, each of which come with a hyperparameter, we also include results for \ac{ig} and Shapley value sampling for comparison. Note that we do not include the Shapley value sampling method in the ImageNet experiments for computational reasons.  Results of our comparison are shown in Fig.\ \ref{fig:comparison_vgg16_g16_all} and Fig.\ \ref{fig:comparison_d3_g14_all}. Exact numerical values are reported in Appendix \ref{app:sweets-spots}.

\begin{figure}[ht!]
    \centering
    \includegraphics[width=0.999\textwidth]{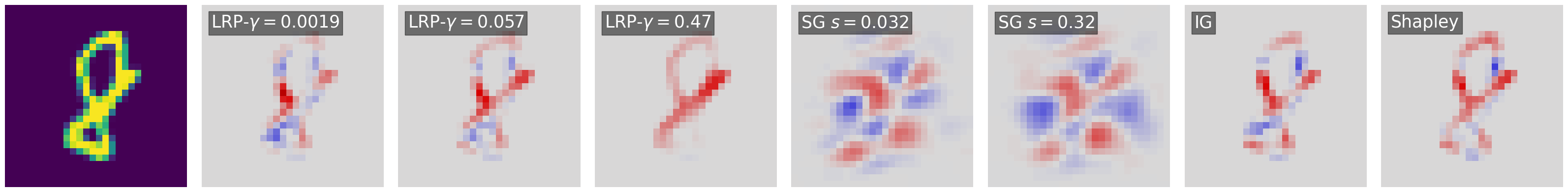}
    \includegraphics[width=0.999\textwidth]{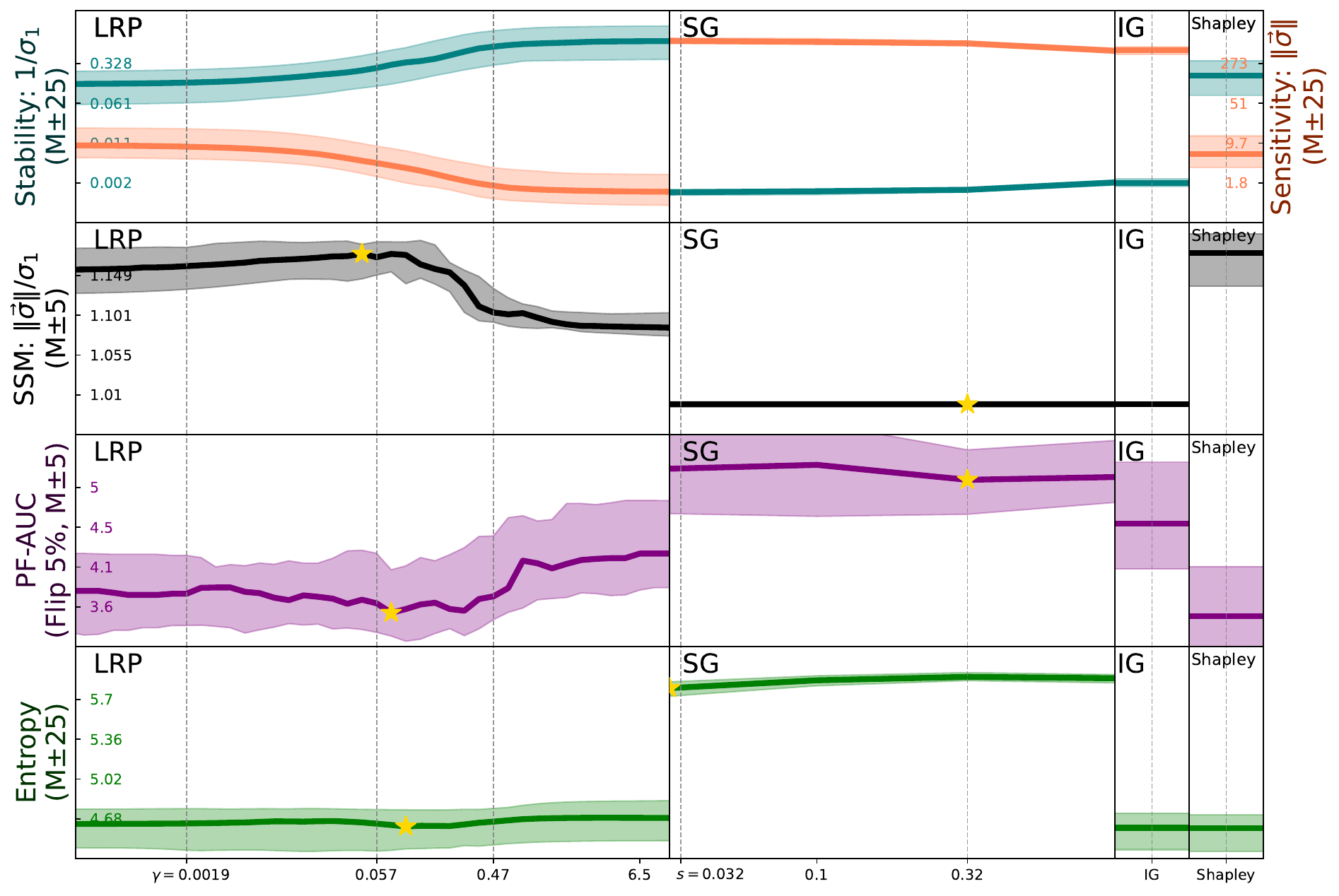}
    \caption{
Quantitative analysis of evaluation metrics for explanation methods on an MNIST-trained model using  100 images. The top row shows exemplary heat maps for the class `8’, one per method and parameter choice per dashed vertical line drawn in the main plot. The structure of this figure follows Fig. \ref{fig:comparison_vgg16_g16_all}, with the following difference: The rightmost column includes results for the Shapley methods. The LRP curve was obtained using  80 different $\gamma$ parameters.
}
    \label{fig:comparison_d3_g14_all}
\end{figure}

For the LRP method, we observe in both models that as the $\gamma$ parameter increases, the explanation \emph{stability} increases while its \emph{sensitivity} decreases. Our analysis suggests the existence of a `sweet spot', where \emph{stability} and \emph{sensitivity} can be both achieved, as indicated by a high value for the sensitivity-stability metric (SSM). In the LRP case, this sweet spot corresponds to choosing an intermediate value of the $\gamma$ parameter ($\gamma=0.04$ for ImageNet and MNIST, cf.\ Appendix \ref{app:sweets-spots}). For the \ac{sg} explanation method, we observe a trend similar to LRP, where increasing the smoothing hyperparameter $s$ results in higher \emph{stability} but lower \emph{sensitivity}. For the ImageNet model, SSM exhibits a clear preference for a specific parameter, whereas on MNIST, the smoothing hyperparameter seems to have no effect on SSM.

Comparing the result of our stability-sensitivity analysis with the pixel-flipping evaluation, specifically, the PF-AUC score described above, we see that both methods reveal a preference for intermediate values of $\gamma$. 
Extending the comparison to the entropy-based evaluation, we see that the \emph{entropy} expresses a similar preference for intermediate values of the $\gamma$ parameter in LRP. This close relation between SSM and \emph{entropy} is particularly intriguing as the \emph{entropy} was originally intended as a test for the presence of noise in the explanation, in other words, a measure of \textit{stability} only. It turns out that the \emph{entropy} metric does more than that: Stable but insensitive explanations are also highly entropic due to the spreading of relevance scores onto excessively many pixels. Thus, our analysis suggests that \emph{entropy} is a fairly holistic measure of explanation quality.

Furthermore, we observe that entropy fails in some cases in its original aim to characterize explanation stability (i.e.\ absence of noise). The \ac{sg} explanation with low smoothing, which is objectively highly instable, appears not to be so when looking at its low entropy score. This discrepancy can be traced to the normalization step before the entropy computation, which tends to ignore the magnitude of noise patterns, causing noise to be neglected when it is collocated with very strong noise occurring only on few pixels. 

Overall, our analysis has revealed that the SSM metric and most evaluation methods witness a subtle interplay between different factors of explanation quality such as stability and sensitivity. Explanation hyperparameters are shown to be effective in influencing those factors of explanation quality. However, these evaluation methods disagree on what precise hyperparameter values are optimal. This imposes caution in drawing general conclusions from the results of specific explanation evaluations. 

\subsection{Expanded Explanations with Spectral Analysis}

\begin{figure}[h]
    \centering \label{fig:paddle_scvec_hms}
    \includegraphics[width=0.99\textwidth]{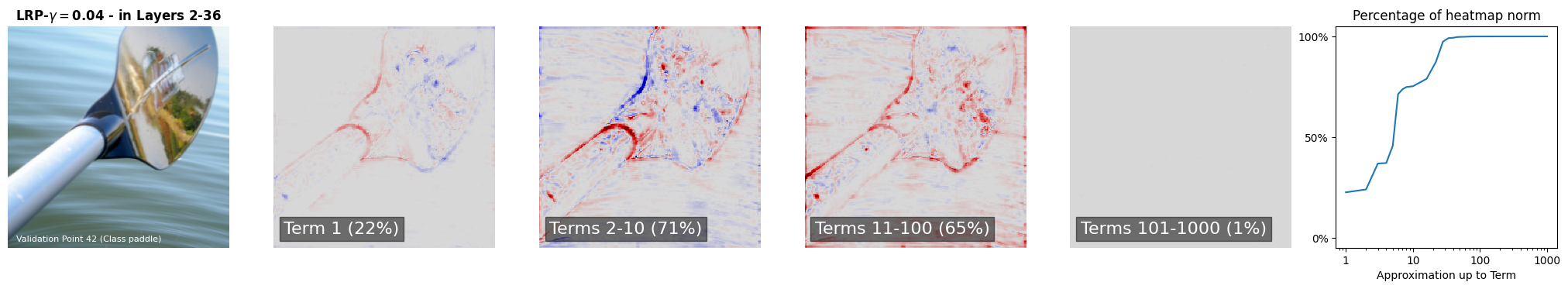}
    \includegraphics[width=0.99\textwidth]{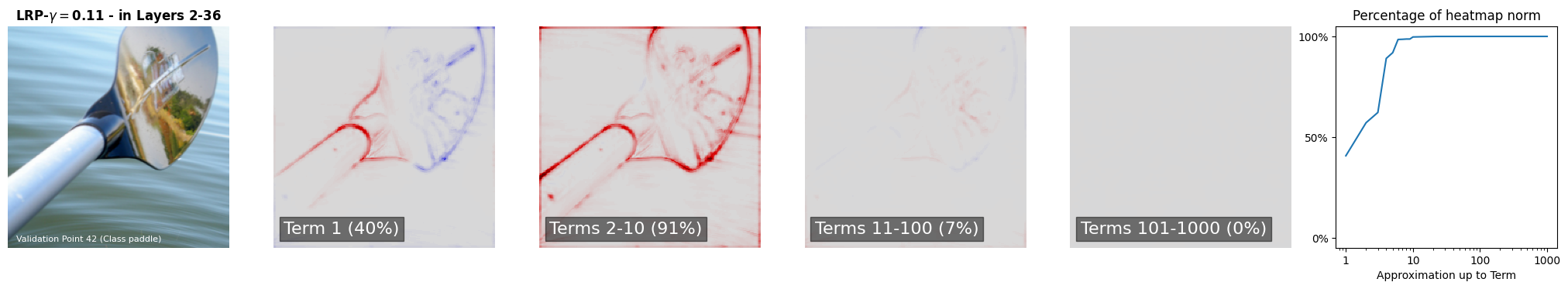}
    \caption{
    LRP explanations for the class `paddle' decomposed into contributions of different singular values (cf.\ Eq.\ \eqref{eq:expanded}). 
    Heat maps visualize how bins of singular values, namely, the ranges $(1,1)$, $(2,10)$, $(11,100)$, and $(101,1000)$ contribute, with the norm of this partial result (as a percentage of the norm of the full heat map) denoted in brackets. The top row of heat maps corresponds to LRP with $ \gamma=0.04 $ and the bottom row uses $ \gamma=0.11 $. The plots on the right visualize the rise in the heat maps norm as it is produced with approximations of $ R_{\cdot| \cdot} $ with increasing rank $k$: $\|\sum_{i=1}^{k} \mathcal{E}(y; \sigma_i)\|_2 \cdot \| \mathcal{E}(y) \|_2^{-1} $. Depending on the choice of $\gamma$, small singular values contribute little to heat maps and their norm, indicating that explanations are sensitive to only a low number of patterns in the data.
    }
    \label{fig:svecs_hms}
\end{figure}

In this section, we show another use of our spectral analysis, which follows from the decomposition of the explanations it offers in terms of singular values; in particular, we recall that the explanation $\mathcal{E}(y)$ can be rewritten as:
\begin{align}
\mathcal{E}(y) & = \sum_{i=1}^K \underbrace{\sigma_i u_i v_i^\top y}_{\displaystyle \mathcal{E}(y;\, \sigma_i)}
\label{eq:expanded}
\end{align}
Each term $\mathcal{E}(y;\, \sigma_i)$ of the sum has the same shape as the original $\mathcal{E}(y)$ and can therefore also be rendered as a heat map. Furthermore, the collection of heat maps sums to the original heat map and can thus be seen as a sum-decomposition of the original explanation. This analysis is shown in Fig.\ \ref{fig:svecs_hms} for an image predicted by the same model, but explained by two versions of LRP (with different $\gamma$ parameters). 

As the $\gamma$ parameter increases, the bulk of the explanation is shifted to a smaller number of leading singular values. These are also associated with less noisy singular vectors. As a result, this shift--associated with an increase in $\gamma$--causes both a denoising of the explanations and the effective degree of freedom in which explanations vary to drop.

\FloatBarrier

%% file: chapters/07conclusionandfuture.tex
\section{Conclusion}
\label{sec:conclusion}

XAI was originally conceived to increase the transparency of complex, nonlinear machine learning methods. Over time, a wide array of evaluation metrics has emerged to assess the quality of explanations, yet this diversity has created new challenges in determining which metric is most appropriate for a particular application and in correctly interpreting the resulting scores. Consequently, it is crucial to develop a better theoretical understanding of the underlying factors that determine explanation quality, as well as how these factors are weighted in different evaluation metrics.

In this work, we propose a novel formal analysis framework that elucidates the multifaceted nature of explanation quality. By applying a spectral analysis of the explanation-generating process, our approach mathematically characterizes two distinct evaluation factors, explanation \emph{stability} and explanation \emph{sensitivity}, which jointly contribute to achieving high-explanation quality.

Moreover, extensive simulations on MNIST and ImageNet models illustrate how the factors of explanation quality align with popular explanation metrics such as pixel-flipping or the explanation's entropy. 

We also demonstrate how to operationalize our framework by decomposing explanations into their spectral components, distinguishing primary explanation factors from secondary effects or noise. Overall, these findings underscore the potential of our conceptual framework to guide the search for better, more consistent explanations.

%% file: chapters/11appendix.tex
\section{Numerical results of explanation quality}
\label{app:sweets-spots}

Table \ref{tab:sweetspots} provides numerical values for the stability-sensitivity metric (SSM), Pixel Flipping (PF-AUC), and Shannon Entropy metric displayed in Figures \ref{fig:comparison_vgg16_g16_all} and \ref{fig:comparison_d3_g14_all}.
On the \LG and \ac{sg} methods, we report the values for the optimal parameter choice $\gamma$ or $\sigma$ (shown in brackets).

\begin{table}[h]
        \caption{
        Explanation quality metrics for the ImageNet and MNIST datasets. We report the median score over 100 validation images.
        The values correspond to the best parameter setting for each method; parameter choices for \LG and \SG are denoted in brackets. ($\uparrow$) The higher the better, and ($\downarrow$) the lower the better.
    }
    \label{tab:sweetspots}
    
    \centering \medskip
    \setlength{\tabcolsep}{5pt}
    \begin{tabular}{rc c c c}
        \toprule
        & LRP ($\gamma$) & SG ($s$) & IG & Shapley \\
        \midrule
        \multicolumn{1}{l}{\textit{ImageNet}}\\[-1mm]
        SSM ($\uparrow$)~~ & $1.24$ ($0.04$)   & $1.31$ ($0.32$)       & $1.26$ & --- \\
        PF-AUC (5\%) ($\downarrow$)~~ & $4.81$ ($0.3$)  & $4.81$  ($0.1$)    & $5.03$ & --- \\
        Shannon Entropy ($\downarrow$)~~ & $10.99$ ($0.30$)    & $11.38$ ($0.03$) &  $11.14$ & --- \\[2mm]

        \multicolumn{1}{l}{\textit{MNIST}} \\[-1mm]
        SSM ($\uparrow$)~~ & $1.18$ ($0.04$)  & $1.00$ ($0.32$)     & $1.00$ & $1.18$ \\
        PF-AUC (5\%) ($\downarrow$)~~ & $3.53$ ($0.07$)  & $5.10$ ($0.32$) &  $4.58$ & $3.49$ \\
        Shannon Entropy ($\downarrow$)~~ & $4.61$ ($0.10$)  & $5.80$ ($0.03$) & $4.60$ & $4.60$ \\
        \bottomrule
    \end{tabular}
\end{table}

\section{Analytical form of LRP-$\gamma$ operator norms}

The LRP-$\gamma$ rule leverages the preference towards positive contributions in reassigning relevance as a stabilizing effect in the creation of heat maps \cite{montavon2019layer}.
We now provide an analytical perspective on this heuristic, demonstrating that $\gamma$ has a domain-specific regularization effect on the explanation process, reducing the operator norm of the conditional relevance matrix $R_{\cdot|\cdot}$.
In this section, we define $p_k := \sum_j [a_j w_{j,k}]^+$ and $n_k := \sum_j [a_j w_{j,k}]^-$ as the sum of all positive or negative inputs to the neuron $k$. We then focus on ReLU-activated neurons, denoting their activation value by $0 \leq |n_k| < p_k$. 

We first observe that every element of the conditional relevance matrix decreases monotonically in magnitude as $\gamma$ increases, ultimately favoring smaller outputs in the matrix multiplication.
Negative entries in the matrix have $R_{j|k} 
= { [a_j w_{j,k}]^-
}/({
(1 + \gamma)p_{k} + n_{k}
})$;
the conditional relevance goes towards 0 because the denominator increases with $\gamma$.
Positive entries of the matrix have $R_{j|k}
= {(1 + \gamma)[a_j w_{j,k}]^+
}/({
(1 + \gamma)p_{k} + n_{k}
})$
and the derivative
${\partial R_{j|k}(\gamma)}/{\partial \gamma}
=
{
[a_j w_{j,k}]^+ \cdot n_k
}/{((1 + \gamma)p_{k} + n_{k})^2}
< 0$ is negative (because $n_k < 0$).
The conditional relevance decrease monotonically towards
$\lim_{\gamma \rarr \infin} R_{j|k} = \frac{ a_j [w_{j,k}]^+ }{p_{k}}$.

Moreover, we establish a precise relationship between the L1 operator norm of LRP explanation steps and the $\gamma$ parameter.
The key observation is that LRP normalizes the total relevance leaving each neuron:
while both positive and negative values of any magnitude are allowed, they must cancel each other so that each column sums to 1, c.f. \Cref{sec:relate_work_eval_method}.
This requirement aligns with the L1 operator norm's role of measuring the sum of all \textit{absolute} relevances going out of a neuron.
Leveraging these parallels, we derive an analytical expression for the L1 operator norm in terms of $\gamma$.

\textbf{First, the L1 norm of a column} $k$ of the conditional relevance matrix is:

\begin{align*}
\| R_{\cdot|k}(\gamma) \|_1 
= \sum_{j} | R_{j | k}(\gamma) |
&= \sum_{j} | 
\frac {
(1 + \gamma)a_j [w_{j,k}]^+ + a_j [w_{j,k}]^-
}{
(1 + \gamma)p_{k} + n_{k}
}|
\\
&= \frac {
\sum_{j} | 
(1 + \gamma)a_j [w_{j,k}]^+ + a_j [w_{j,k}]^-
| }{
(1 + \gamma)p_{k} + n_{k}
}
\end{align*}

Then, using the fact that $([a_j w_{j,k}]^+ = 0) \vee ([a_j w_{j,k}]^-=0)$ for a given $(j,k)$ we can pull the absolute value operation into the terms of the denominator and use our definition of the summed negative and positive contributions to a neuron as $p_k$ and $n_k$:

\begin{align*}
\| R_{\cdot|k}(\gamma) \|_1 
&= \frac {
\sum_{j} 
|(1 + \gamma)[a_j w_{j,k}]^+| + |[a_j w_{j,k}]^-|
}{(1 + \gamma)p_k + n_k}
\\
&= \frac {
|\sum_{j} (1 + \gamma)[a_j w_{j,k}]^+| + |\sum_{j} [a_j w_{j,k}]^-|
}{(1 + \gamma)p_k + n_k}
\\
&= \frac {(1 + \gamma)p_k + |n_k|}{(1 + \gamma)p_k - |n_k|}
= 1 + \frac{2|n_k|}{(1 + \gamma)p_k - |n_k|}
= 1 + \frac{2 \frac{|n_k|}{p_k}}{1 - \frac{|n_k|}{p_k} + \gamma}
\end{align*}

By defining $c_k := \frac{|n_k|}{p_k}$, we obtain a compact expression for the L1 column norm in dependence of $\gamma$:

\begin{align*}
\| R_{\cdot|k}(\gamma) \|_1 
= 
1 + \frac{2 c_k}{1 - c_k + \gamma}
\end{align*}

\textbf{The induced L1 operator norm} is then simply the maximum over all columns $k'$ norms.
While all column norms decrease with $\gamma$, the column index $k$ of the column that is largest stays constant across the whole domain of $\gamma$.
This allows us to define a constant ‘coefficient’ $c$ for the entire operator and write the operator norm in a compact form. With $c := \max_{k', |n_{k'}|<p_{k'}} \frac{|n_{k'}|}{p_{k'}}$ and $k$ as the $argmax$ of the expression,

\begin{align}
\label{eq:l1}
\| R_{\cdot|\cdot}(\gamma)  \|_1  
= \max _{k'} \| R_{\cdot|k'}(\gamma) \|_1 
= \| R_{\cdot|k}(\gamma) \|_1 = 1 + \frac{2 c}{1 - c + \gamma}.
\end{align}

When $\gamma = 0$, the L1 norm $\| R_{\cdot|\cdot}(0) \|_1$ can become large if there is an activated neuron $k$ in the layer, whose negative inputs are almost as large as it’s negative input: $|n_k| \lessapprox p_k \Rarr c \lessapprox 1$.
The LRP-$\gamma$ rule outweighs the positive contributions when assigning relevance. If the positive inputs already dominate in every neuron, the operator norm is not significantly changed by increasing the $\gamma$-parameter.

Finally, we can \textbf{bound the operator norms of the entire explanation method} as a function of the LRP-$\gamma$ parameters in the layers that use the rule.
Simplifying to the case that every layer $0 \leq t < T$ of the network uses the LRP-$\gamma$ rule, and that the measured $c^{(t)}$ is the same for all layers $t$, we find that the operator norm of the entire explanation process is bounded by
\begin{align}
    \notag
    \| A^{(0\larr T)}_{\cdot|\cdot}(\gamma) \|_1
    &\leq
    \prod_{t=0}^{T} \| A^{(t)}_{\cdot|\cdot} (\gamma) \|_1
    =
    \big(1 + \frac{2 c}{1 - c + \gamma}\big)^T
    =
    \sum_{t=0}^T
    \binom{T}{t}
    \big(\frac{2 c}{1 - c + \gamma}\big)^T
    \label{eq:l1_norm_deep}
\end{align}

The operator norm bound decreases monotonically with $\gamma$.
In fact, for large $c\approx 1$, the bound can shrink rapidly, at a rate up to $\gamma^{-T}$.
Practitioners commonly observe this effect in deep networks explained with the \LG rule across multiple consecutive layers (e.g.\cite{pahde2023optimizing}, \cite{montavon2019layer}).
Specifying too high $\gamma$ quickly leads to very high \textit{Stability} and low \textit{Sensitivity}.
The explanations have a coarse-grained nature and can not distinguish between classes anymore, as visualized in \Cref{fig:paddle_scvec_hms} \textit{(bottom)}.

\section{Computational Complexity of the Evaluation Methods}
\label{sec:computational_complexity}

\textit{Stability}, as the largest singular value, can be computed efficiently using iterative methods such as the power iteration or Lanczos algorithms \cite{Lanczos1950}. Assuming a network with $d$ inputs and $h$ outputs, and its redistribution matrices as described in Section \ref{section:method}, these methods have a computational cost of  $\mathcal{O}(dh)$  per iteration. The number of iterations until the first singular value and vector are found depend on the desired precision and the spectral gap of the matrix.
\textit{Sensitivity} can be calculated as the square root of the sum of the squared entries of the redistribution matrix, which is equivalent to its Frobenius norm. The computational complexity for this calculation is $\mathcal{O}(dh)$.

\paragraph{Complexity of constructing the redistribution matrix.}
In perturbation-based methods (such as Shapley), heatmaps for all classes--and thus the entire redistribution matrix--can be created simultaneously by collecting each class's output for every perturbed input. Therefore the complexity to create the redistribution matrix is equivalent to the complexity of the explanation for one target~output.

Propagation-based methods such as LRP and SmoothGrad require both forward and backward passes per (perturbed) input to generate a heatmap. However, the forward computation can be reused to compute gradients for multiple classes. In summary, LRP scales with $\mathcal{O}(h)$ per input image, while SmoothGrad with $k$ noise perturbations scales with $\mathcal{O}(hk)$.

\section{MNIST CNN architecture}
\label{sec:mnist_cnn_arch}

The convolutional neural network (CNN) architecture used for the MNIST dataset is a small network with 6 convolutional layers containing ReLU activation functions.
The full sequence includes 3 convolutional layers with kernel sizes $3\times3$, $3\times3$, and $5\times5$ using $8$ filters each, followed by a max-pooling layer with kernel size $2$.
This is followed by convolutional layers of kernel sizes $5\times5$, $3\times3$, and $3\times3$ and $16$ filters each.
After a second max-pooling layer (with a kernel size of 2), the output is flattened and passed through a fully connected layer to compute the 10 logit scores.
The network is trained using the SGD optimizer and a learning rate of 0.1 until convergence.

The data from the Kaggle Digit Recognizer dataset \cite{kaggle_digit_recognizer} is randomly split into a training and test set with a ratio of $80\%$ to $20\%$.
The network is trained on the 33,600 training data points using the SGD optimizer and a learning rate of 0.1 until convergence.
100 images are drawn from the test set for the evaluation of the explanation methods.